\documentclass[10pt,letterpaper,twocolumn]{article}
\usepackage[latin9]{inputenc}
\usepackage{array}
\usepackage{multirow}
\usepackage{amsmath}
\usepackage{amssymb}
\usepackage{graphicx}

\makeatletter

\pdfpageheight\paperheight
\pdfpagewidth\paperwidth

\providecommand{\tabularnewline}{\\}

\usepackage{cvpr}\usepackage{times}\usepackage{epsfig}

\usepackage[pagebackref=true,breaklinks=true,letterpaper=true,colorlinks,bookmarks=false]{hyperref}

\cvprfinalcopy % *** Uncomment this line for the final submission

\ifcvprfinal\fi
\usepackage{tablefootnote}
\usepackage{threeparttable}

\makeatother

\begin{document}

\title{Scene recognition with CNNs: objects, scales and dataset bias}

\author{Luis Herranz, Shuqiang Jiang, Xiangyang Li\\
Key Laboratory of Intelligent Information Processing of Chinese Academy
of Sciences (CAS)\\
Institute of Computer Technology, CAS, Beijing, 100190, China\\
\texttt{\small{}\{luis.herranz,shuqiang.jiang,xiangyang.li\}@vipl.ict.ac.cn}}
\maketitle
\begin{abstract}
Since scenes are composed in part of objects, accurate recognition
of scenes requires knowledge about both scenes and objects. In this
paper we address two related problems: 1) scale induced dataset bias
in multi-scale convolutional neural network (CNN) architectures, and
2) how to combine effectively scene-centric and object-centric knowledge
(i.e. Places and ImageNet) in CNNs. An earlier attempt, Hybrid-CNN\cite{Zhou2014b},
showed that incorporating ImageNet did not help much. Here we propose
an alternative method taking the scale into account, resulting in
significant recognition gains. By analyzing the response of ImageNet-CNNs
and Places-CNNs at different scales we find that both operate in different
scale ranges, so using the same network for all the scales induces
dataset bias resulting in limited performance. Thus, adapting the
feature extractor to each particular scale (i.e. scale-specific CNNs)
is crucial to improve recognition, since the objects in the scenes
have their specific range of scales. Experimental results show that
the recognition accuracy highly depends on the scale, and that simple
yet carefully chosen multi-scale combinations of ImageNet-CNNs and
Places-CNNs, can push the state-of-the-art recognition accuracy in
SUN397 up to 66.26\% (and even 70.17\% with deeper architectures,
comparable to human performance).
\end{abstract}

\section{Introduction}

State-of-the-art in visual recognition is based on the successful
combination of deep representations and massive datasets. Deep convolutional
neural networks (CNNs) trained on ImageNet (i.e. ImageNet-CNNs) achieve
impressive performance in object recognition, while CNNs trained on
Places (Places-CNNs) do in scene recognition\cite{Zhou2014b,Jeff2014}.
However, CNNs also have limitations, such as the lack of invariance
to significant scaling. This problem is particularly important in
scene recognition, due to a wider range of scales and a larger amount
of objects per image.

As an alternative to Places-CNN holistic representation, some recent
works\cite{Gong2014,Dixit2015,Yoo2015,Wu2015} have shown that CNN
features extracted locally in patches can be also aggregated into
effective scene representations. Often, these approaches combine multiple
scales, that are pooled using VLAD\cite{Gong2014} or Fisher vector
(FV)\cite{Yoo2015} encoding. Dixit et al\cite{Dixit2015} suggested
applying the pooling directly on the semantic representation, arguing
that semantic representations are more invariant. Recently, Wu et
al\cite{Wu2015} proposed an architecture in which dense sampling
of patches is replaced by region proposals and discrimintive patch
mining. In general, these works use ImageNet-CNN to extract the local
activations instead of Places-CNN, since local patches are closer
to objects than to scenes. However, a largely overlooked aspect in
this multi-scale scenario is the role of the scale and its relation
with the feature extractor (i.e. CNN). One limitation of current multi-scale
approaches is the naive use of CNNs by simply considering CNNs as
general purpose feature extractors\cite{Razavian2014,Jeff2014}. Using
the same fixed CNN model for all the scales inevitably leads to dataset
bias\cite{Torralba2011}, since the properties of the data vary at
different scales, while the feature extractor remains fixed.

Since objects are main components of scenes, knowledge about objects
may be helpful in scene recognition. Although, Places-CNN itself develops
suitable object models at intermediate layers\cite{Zhou2015}, the
information in ImageNet might be valuable. However, in a previous
attempt, a network trained with the combined dataset ImageNet+Places
(Hybrid-CNN\cite{Zhou2014b}) show that including ImageNet, far from
helpful was harmful. We will see how this problem is also connected
to scale-related dataset bias.

In this paper we will study these two problems (i.e. dataset bias
in patch-based CNNs under different scaling conditions, and how to
effectively combine Places and ImageNet) and will see that they are
related. Torralba and Efros\cite{Torralba2011} studied the dataset
bias as a cross-dataset generalization problem, in which the same
classes may have slightly different feature distributions in different
datasets. In our particular case, this bias in the feature distribution
is induced by scaling the image. If the scaling operation is considerable,
the characteristics of the data may change completely, switching from
scene data to object data. Understanding and quantifying this bias
can help us to design better multi-scale architectures, and even better
ways to combine object and scene knowledge. In particular, we propose
multi-scale architectures with \textit{scale-specific networks} as
a principled way to address scale-related dataset bias and combine
scene and object knowledge (i.e. Places and ImageNet). In the next
sections:
\begin{itemize}
\item We show that using a single CNN as a generic feature extractor from
patches is quite limited, due to the dataset bias induced by scale
changes. We show how networks trained on different datasets are suitable
for different scale ranges. In particular, ImageNet-CNN and Places-CNN
have very different optimal ranges, due to their object-centric and
scene-centric natures.
\item We evaluate two strategies to alleviate the dataset bias by using
scale-specific networks: hybrid Places/ImageNet architectures and
fine tuning. By combining after reducing the dataset bias, our method
is also a more effective way to combine Places and ImageNet. Extensive
experiments with different scale combinations and hybrid variations
(optionally fine tuned) lead to some variations achieving state-of-the-art
performance in scene recognition. 
\end{itemize}

\section{Objects and scenes}

\subsection{Objects in object datasets and scene datasets}

The knowledge learned by CNNs lies in the data seen during training,
and will be of limited use if tested in a different type of data.
Thus, CNNs trained with ImageNet are limited when used for scene recognition
due to this training/test bias, while Places-CNNs perform better in
this task. While this is essentially true, objects and scenes are
closely related, so knowledge about objects can be still helpful to
recognize scenes, if used properly.

Understanding the characteristics of the datasets involved is essential
to better explain the causes of dataset bias. In our case, we want
to analyze the properties of objects found in scene and object datasets.
We focus on two aspects related with the objects: \textit{scales}
and \textit{density}.

To evaluate the dataset bias we use SUN397\cite{Xiao2010a,Xiao2014a}
as target dataset. Since Places contains scene data, with 205 scene
categories overlapping with SUN397, and significantly more data, we
can expect a low dataset bias. Thus we focus on ImageNet (in particular
ILSVRC2012), which contains mostly object data. Fortunately, both
ImageNet and SUN have a fraction of images with region annotations
and labels, so we can collect some relevant statistics and compare
their distributions (we used the LabelMe toolbox\cite{Torralba2010}).
Since we will use this information to interpret the variations in
recognition accuracy in next experiements, we focus on the 397 categories
of the SUN397 benchmark (rather than the 908 categories of the full
SUN database).

\begin{figure}
\centering{}\setlength{\tabcolsep}{0pt}
\renewcommand{\arraystretch}{1}\setlength{\tabcolsep}{0pt}
\renewcommand{\arraystretch}{1}%
\begin{tabular}{cc}
\includegraphics[width=0.4\columnwidth]{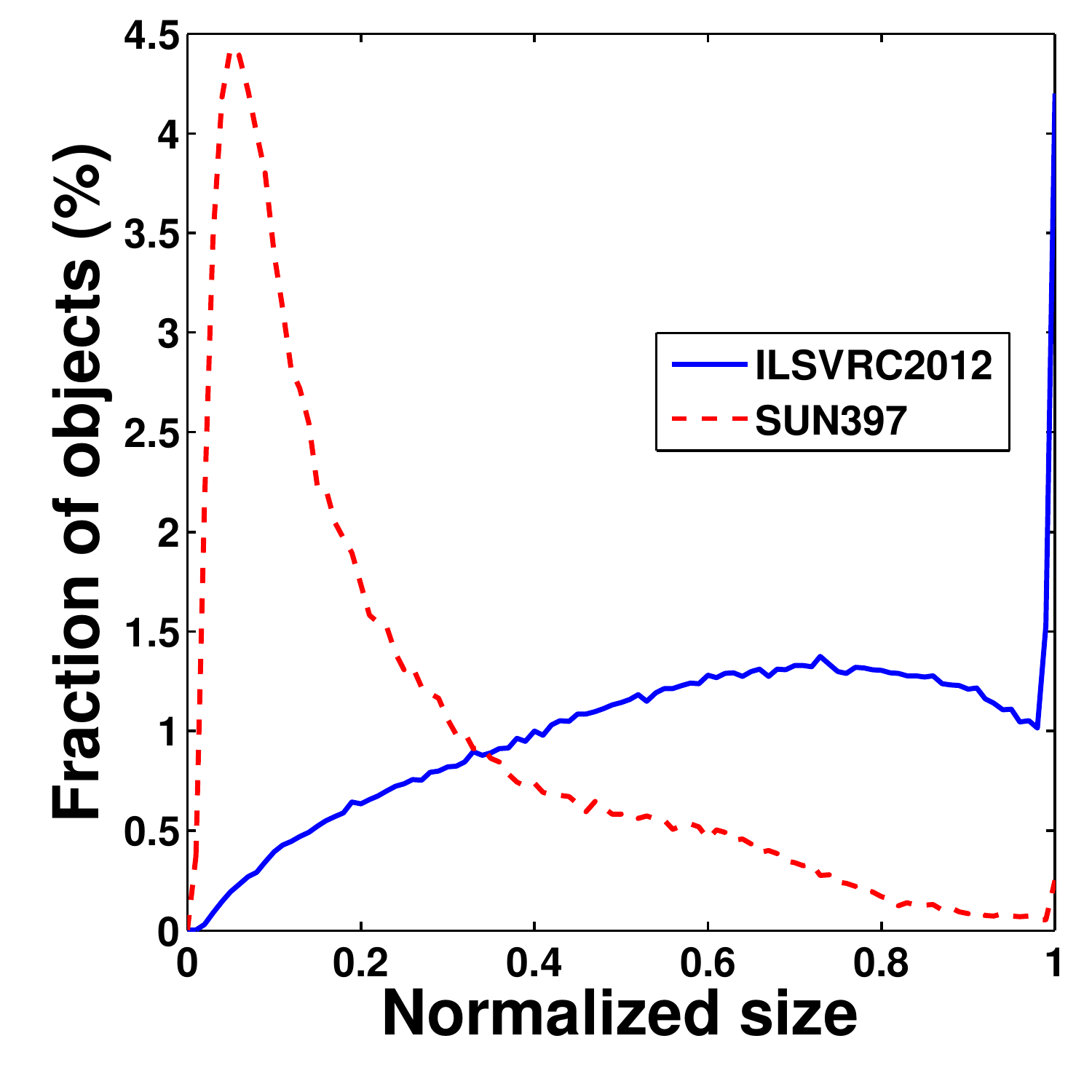} & \includegraphics[width=0.4\columnwidth]{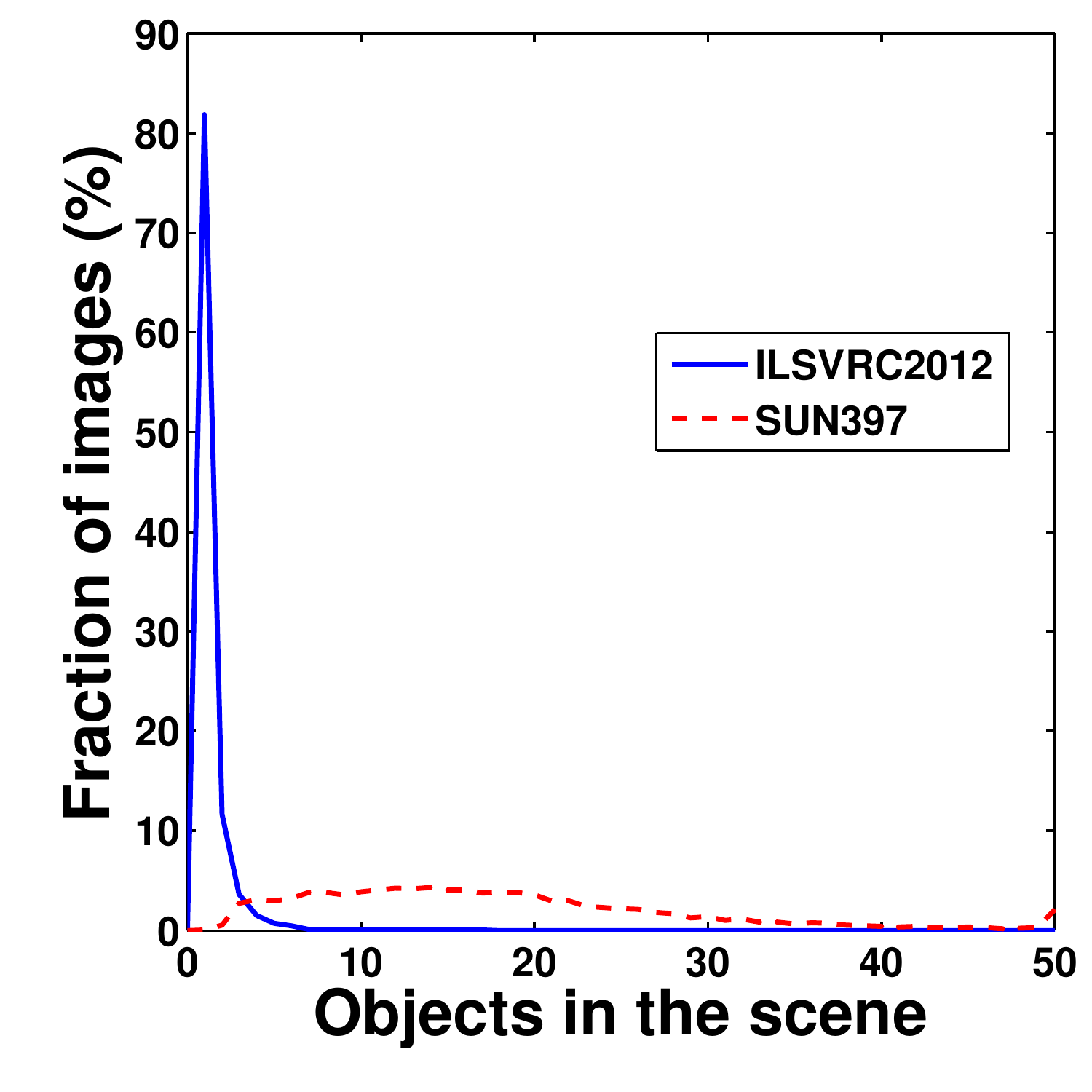}\tabularnewline
(a) & (b)\tabularnewline
\multicolumn{2}{c}{\includegraphics[width=0.9\columnwidth]{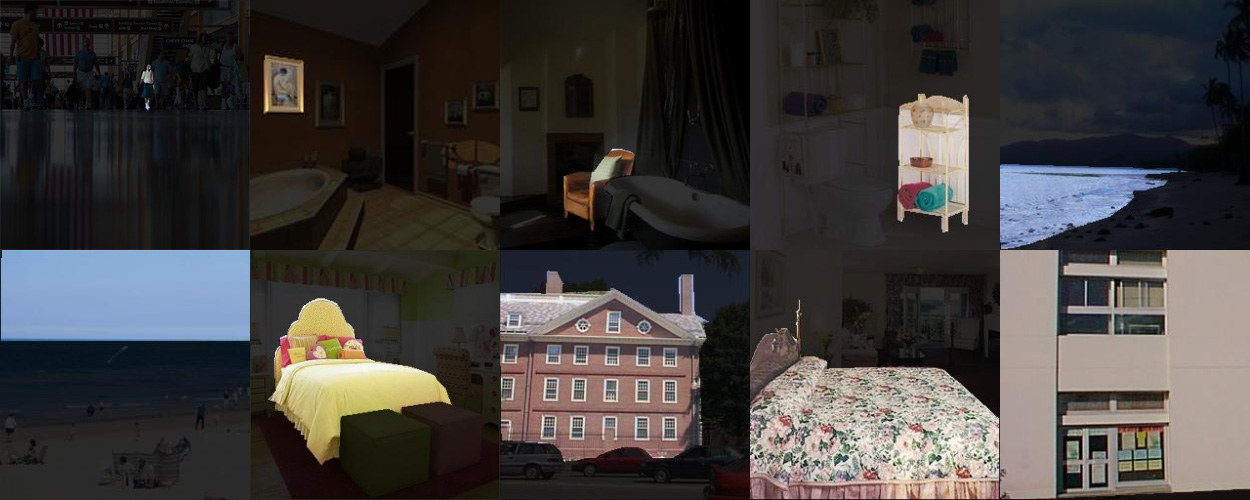}}\tabularnewline
\multicolumn{2}{c}{(c)}\tabularnewline
\end{tabular}\caption{\label{fig:distribution_sizes}Characteristics of objects in ILSVRC2012
(object data) and SUN397 (scene data): (a) distribution of objects
sizes (normalized), (b) number of objects per scene, and (c) examples
of objects by increasing normalized size.}
\vspace{-0.5cm}
\end{figure}

\textbf{Scale.} Fig.~\ref{fig:distribution_sizes}a shows the distribution
of object sizes, and Fig.~\ref{fig:distribution_sizes}c some examples
of objects of different normalized sizes. We normalized the size of
the object relative to the equivalent training crop. While objects
in ImageNet are mostly large, often covering the whole image, objects
in SUN397 are much smaller, corresponding to the real distribution
in scenes. Thus Fig.~\ref{fig:distribution_sizes}a shows an obvious
mismatch between both datasets.

\textbf{Density.} Fig.~\ref{fig:distribution_sizes}b shows the distribution
of object annotations per scene image. We can observe that images
in ImageNet usually contain just one big object, while images in SUN397
typically contain many small objects.

\subsection{Dataset bias in object recognition}

In order to study the behaviour of ImageNet-CNNs and Places-CNNs in
object recognition, we need object data extracted from scenes datasets.
We selected 100 images per category from the 75 most frequent object
categories in SUN397, so we can have enough images to train SVM classifiers.
We took some precautions to avoid selecting too small objects.

\begin{figure*}
\centering{}\setlength{\tabcolsep}{0pt}
\renewcommand{\arraystretch}{1}%
\begin{tabular}{c}
\includegraphics[width=0.8\textwidth]{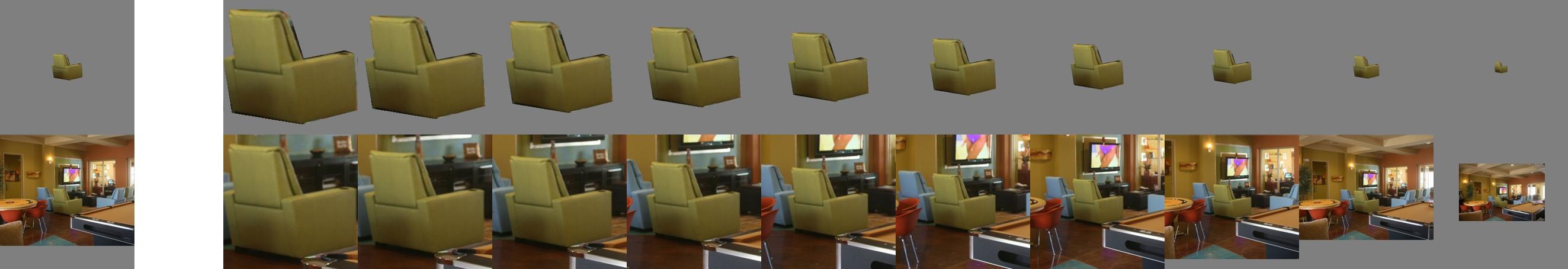}\tabularnewline
(a)\tabularnewline
\includegraphics[width=0.8\textwidth]{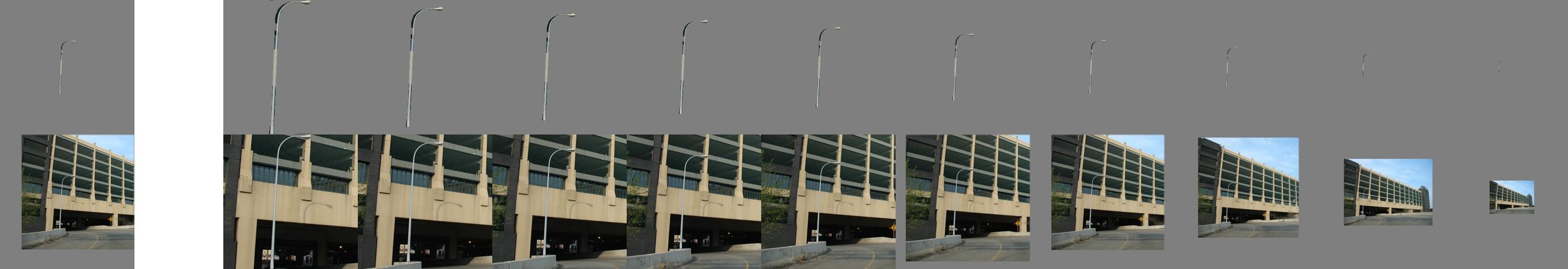}\tabularnewline
(b)\tabularnewline
\end{tabular}\caption{\label{fig:osun75_progressions}The two variants used in the object
recognition experiments: object masked (top row) and object with background
(bottom row) with two examples of (a) \textit{armchair} and (b) \textit{streetlight}.
Left crops show the object in the original scale in the scene. Right
crops show the object scaled progressively from the canonical size
(100\%) down to 10\%. All the images are centered in the object of
interest.}
\end{figure*}

In contrast to most object and scene datasets, in this case we have
the segmentation of the object within the scene, so we can use it
to create variations over the same objects. Thus we defined two scales: 
\begin{itemize}
\item Original scale: the scale of the object in original scene.
\item Canonical scale: the object is centered and rescaled to fill the crop
(keeping the aspect ratio). So in this case its normalized size is
1.
\end{itemize}
Then we created four variations (see Fig.~\ref{fig:osun75_progressions}):
\textit{original masked}, \textit{original with background}, \textit{canonical
masked} and \textit{canonical with background}. In particular, to
study the response to different scaling, the canonical variant is
scaled in the range 10\%-100\%. Note how scaling the variant with
background shifts progressively the content of the crop from object
to scene.

\subsection{Scale sensitivity and object density}

\begin{figure}
\centering{}\setlength{\tabcolsep}{0pt}
\renewcommand{\arraystretch}{1}\includegraphics[width=0.9\columnwidth]{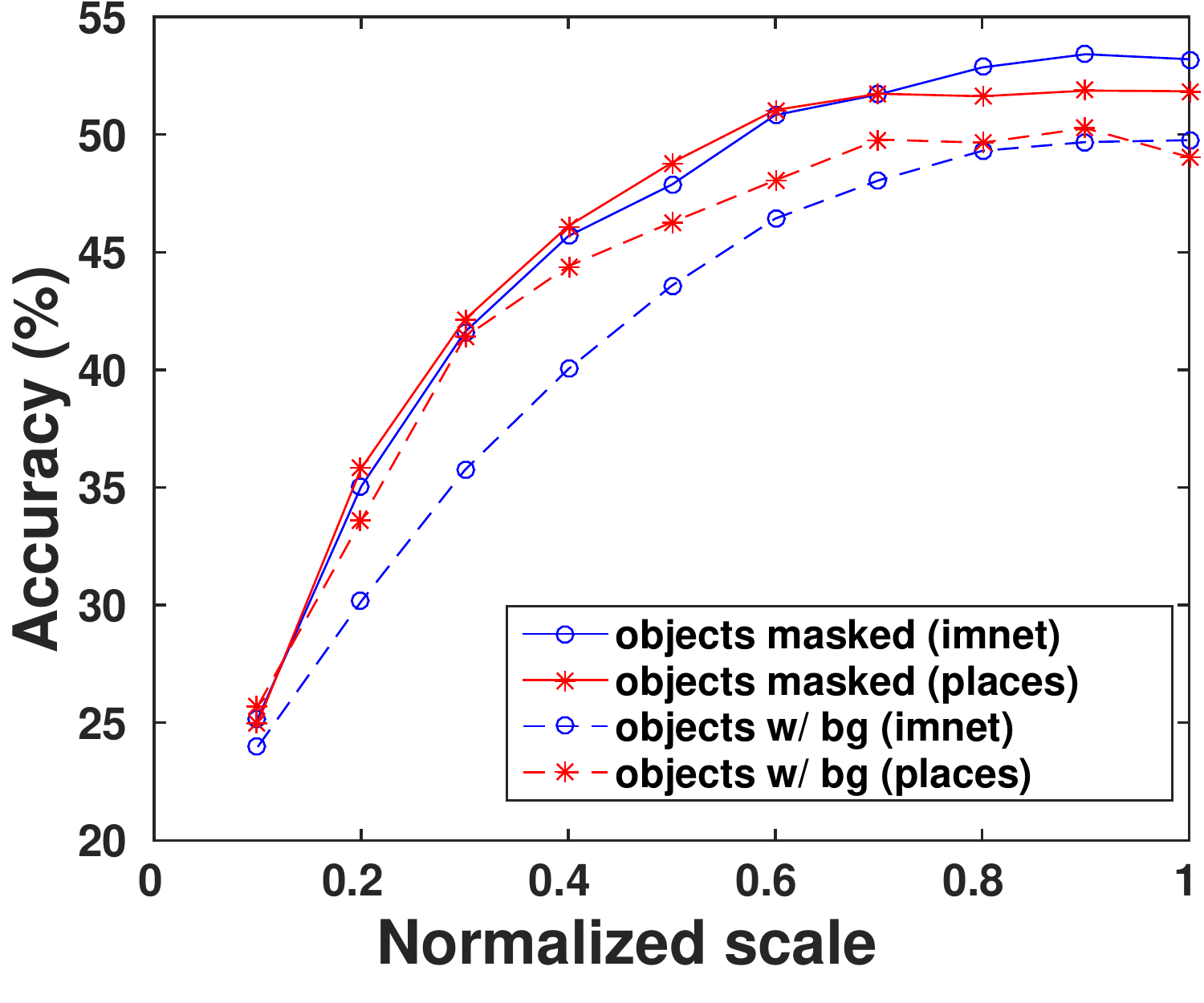}\caption{\label{fig:object_scale_sensitivity}Object recognition accuracy on
SUN397 (75 categories).}
\vspace{-0.3cm}
\end{figure}

We trained a SVM classifier with 50 images per class, and tested on
the remaining 50 images. The input feature was the output of the fc7
activation. The results are shown in Fig.~\ref{fig:object_scale_sensitivity}.
We use two variants: \textit{objects masked} and \textit{objects with
background} (see Fig.~\ref{fig:osun75_progressions}). Regarding
\textit{objects masked}, where the background is removed, we can see
that in general the performance is optimal when the object is near
full size, above 70-80\%. This is actually the most interesting region,
with ImageNet-CNN performing slightly better than Places-CNN. This
is interesting, since Places-CNN was trained with scenes containing
more similar objects to the ones in the test set, while ImageNet-CNN
was trained with the less related categories found in ILSVRC2012 (e.g.
dogs, cats). However, as we saw in Fig.~\ref{fig:distribution_sizes}a,
objects in ILSVRC2012 cover a large portion of the image in contrast
to smaller objects in SUN397, suggesting that a more similar scale
in the training data may be more important than more similar object
categories. As the object becomes smaller, the performance of both
models degrades similarly, again showing a limited robustness to scale
changes.

Focusing now on the \textit{objects with background} variant, the
performance is worse than when the object is isolated from the background.
This behaviour suggests that the background may introduce some noise
in the feature and lead to poorer performance. In the range close
to full object size, both ImageNet-CNN and Places-CNN have similar
performance. However, as the object becomes smaller, and the content
is more similar to scenes, Places-CNN has much better performance
than ImageNet-CNN, arguably due to the fact it has learn contextual
relations between objects and global scene properties. In any case,
scales with low accuracy are probably too noisy and not suitable for
our purpose.

\section{Multi-scale architecture with scale-specific networks}

\subsection{Overview}

\begin{figure*}
\centering{}\includegraphics[bb=0bp 0bp 1509bp 636bp,clip,width=0.9\textwidth]{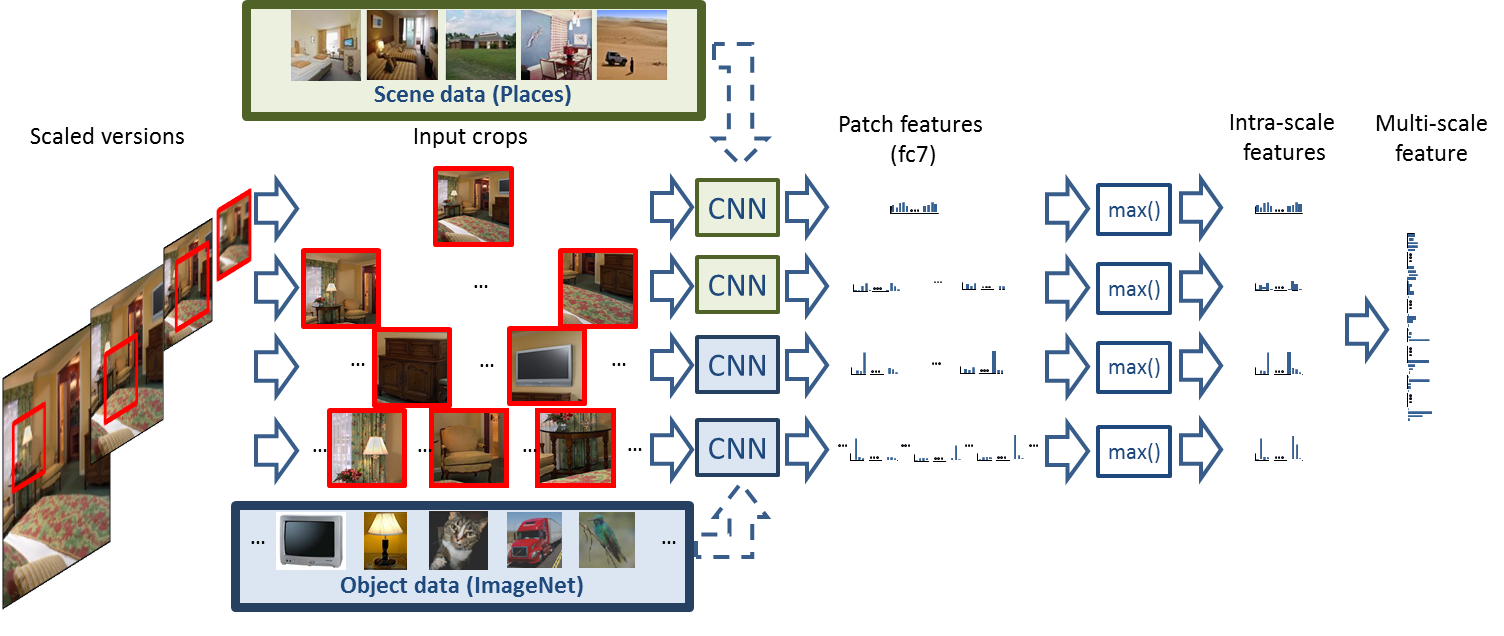}\caption{\label{fig:multiscale_representation}Multi-scale architecture combining
scale-specific networks (spliced architecture). ImageNet-CNNs and
Places-CNNs are combined according to the scale of the input patches.
This can effectively alleviate the dataset bias by adapting test data
to the underlying training data. Intra-scale features are obtained
using max pooling within each scale, and then concatenated into a
single multi-scale feature.}
\vspace{-0.3cm}
\end{figure*}

For scene recognition we introduce our multi-scale architecture, which
combines several networks that operate in parallel over patches extracted
from increasingly larger versions of the input image. We use a standard
multi-scale architecture combining several AlexNet CNNs (Caffe-Net\cite{Jia2014}
in practice) where 227x227 patches are extracted from each full image.
For faster processing, instead of extracting patches independently
we use a fully convolutional network. In contrast to recent works\cite{Gong2014,Gao2015,Yoo2015,Dixit2015},
we adopt simple max pooling to aggregate patch features into image
features.

The previous analysis and experimental results on object recognition
evidence the limitations of using either ImageNet-CNNs or Places-CNNs
to deal with such a broad range of scales, and will be confirmed in
the next sections by the experiments on scene data. For these reasons,
we propose a hybrid architecture introducing two simple, yet crucial
modifications in the architecture (discussed previously in Section~2.2).
\begin{itemize}
\item Instead of using naively the same CNN model for all the scales, we
select the most suitable one for each (ImageNet-CNN, Places-CNN or
fine tuned).
\item Optionally we fine tune each CNN model to further adapt it to the
range of each scale. This requires resizing the image to target size
and extracting patches for training.
\end{itemize}

\subsection{Differences with previous works}

Our architecture is similar to others proposed in previous multi-scale
approaches\cite{Gong2014,Yoo2015,Dixit2015}, with the subtle difference
of using scale-specific networks in a principled way to alleviate
the dataset bias induced by scaling. The main emphasis in these works
is on the way multi-scale features are combined, implemented as either
VLAD or FV encoding, while leaving the CNN model fixed. While adding
a BOW encoding layer can help to alleviate somewhat the dataset bias,
the main problem is still the rigid CNN model. In contrast, our method
addresses better the dataset bias related with scale and achieves
significantly better performance, by simply adapting the CNN model
to the target scale, even without relying to sophisticated pooling
methods.

We can also regard our approach as a way to combine the training data
available in Places and ImageNet. This was explored previously by
Zhou et al\cite{Zhou2014b}, who trained a Hybrid-CNN using the AlexNet
architecture and the combined Places+ImageNet dataset. However, Hybrid-CNN
performs just slightly better than Places-CNN on MIT Indoor 67 and
worse on SUN397. We believe that the main reason was that this way
of combining data from ImageNet and Places ignores the fact that objects
found in both datasets in two different scale ranges (as shown in
Fig.~\ref{fig:distribution_sizes}). In contrast, our architecture
combines the knowledge in a scale-adaptive way via either ImageNet-CNN
or Places-CNN. Wu et al\cite{Wu2015} use Hybrid-CNN on patches at
different scales. Again, the main limitation is that the CNN model
is fixed, not adapting to the scale-dependent distributions of patches.

\section{Experiments on scene recognition}

In this section we perform experiments directly over scene data, to
evaluate the relation beween scale, training dataset and dataset bias
by analyzing the scene recognition performance. Then we combine and
evaluate multi-scale architectures.

\subsection{Datasets}

We evaluate the proposed architectures with three widely used scene
benchmarks. 15 scenes\cite{Lazebnik2006} is a small yet popular dataset
with 15 natural and indoor categories. Models are trained with 100
images per category. MIT Indoor 67\cite{Quattoni2009a} contains 67
categories of indoor images, with 80 images per category available
for training. Indoor scenes tend to be rich in objects, which in general
makes the task more challenging, but also more amenable to architectures
using ImageNet-CNNs on patches. SUN397\cite{Xiao2010a,Xiao2014} is
a larger scene benchmark (at least considered as such before Places)
containing 397 categories, including indoor, man-made and natural
categories. This dataset is very challenging, not only because of
the large number of categories, but also because the more limited
amount of training data (50 images per category) and a much larger
variability in objects and layout properties. It is widely accepted
as the reference benchmark for scene recognition. We consider seven
scales in our experiments, obtained by scaling images between 227x227
and 1827x1827 pixels.

\subsection{Single scale}

\subsubsection{\label{subsec:single_scale_accuracy}Accuracy}

\begin{figure}
\centering{}\setlength{\tabcolsep}{0pt}
\renewcommand{\arraystretch}{1}%
\begin{tabular}{c}
\includegraphics[width=1\columnwidth]{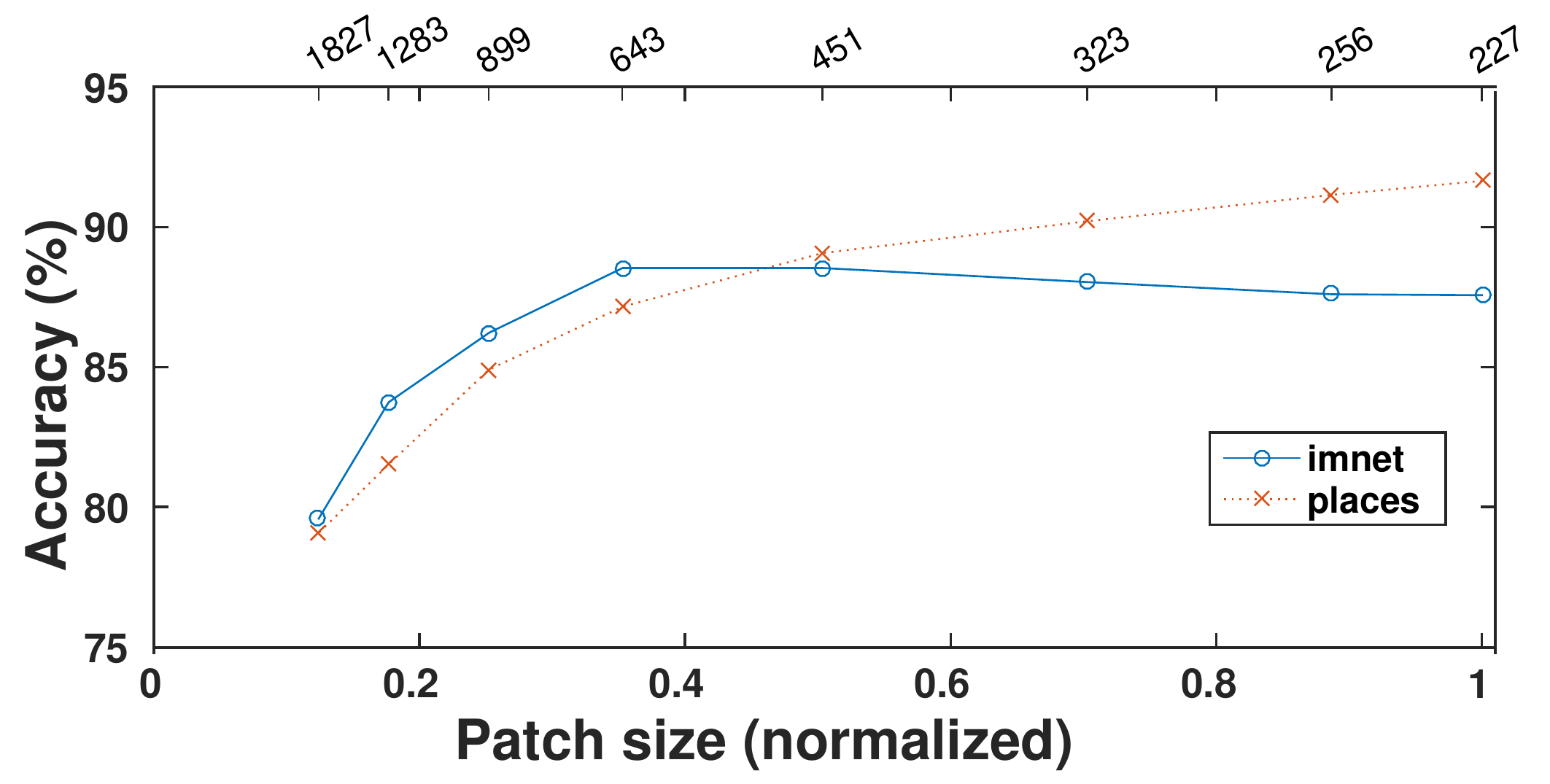}\tabularnewline
(a)\tabularnewline
\includegraphics[width=1\columnwidth]{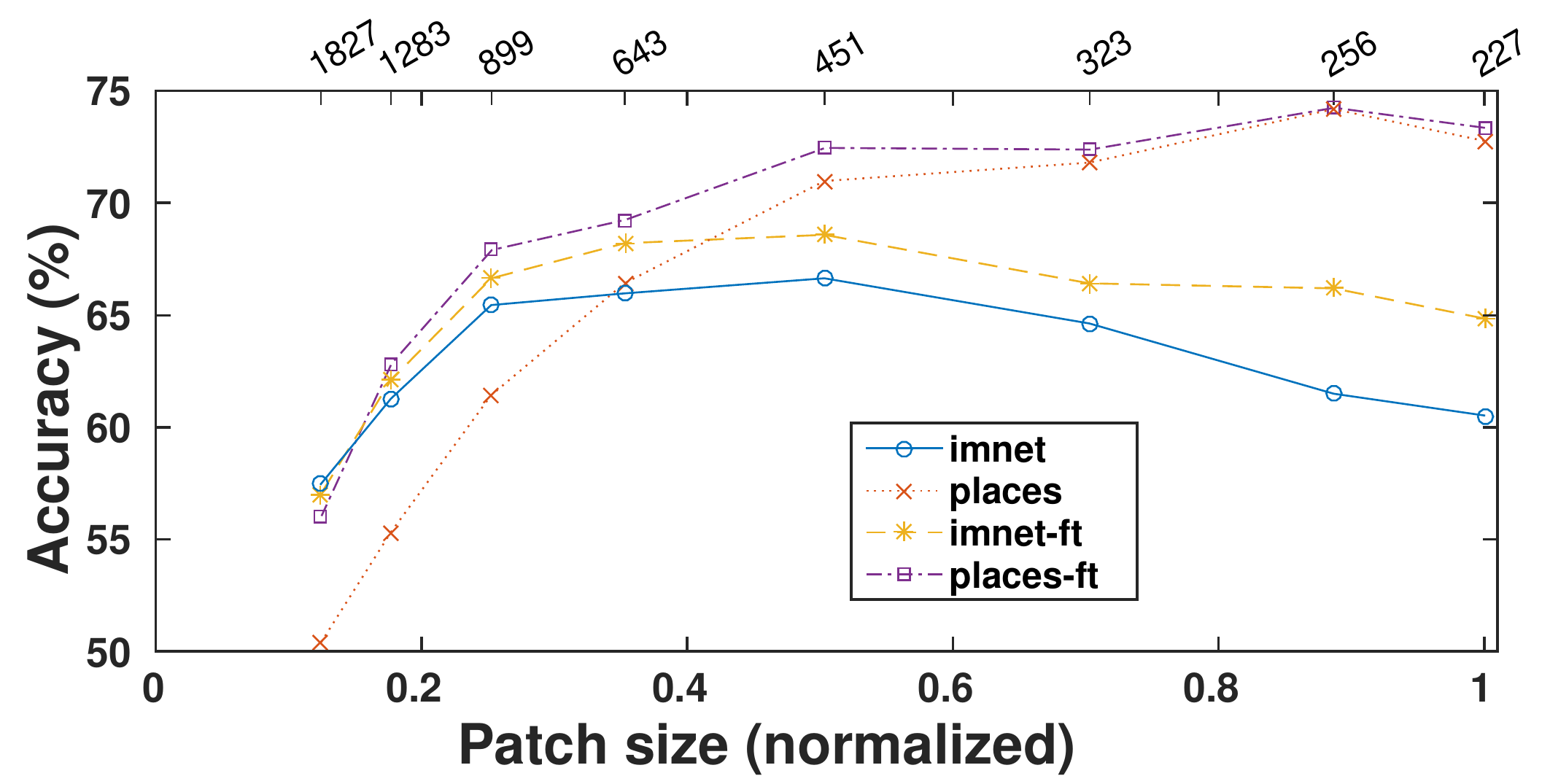}\tabularnewline
(b)\tabularnewline
\includegraphics[width=1\columnwidth]{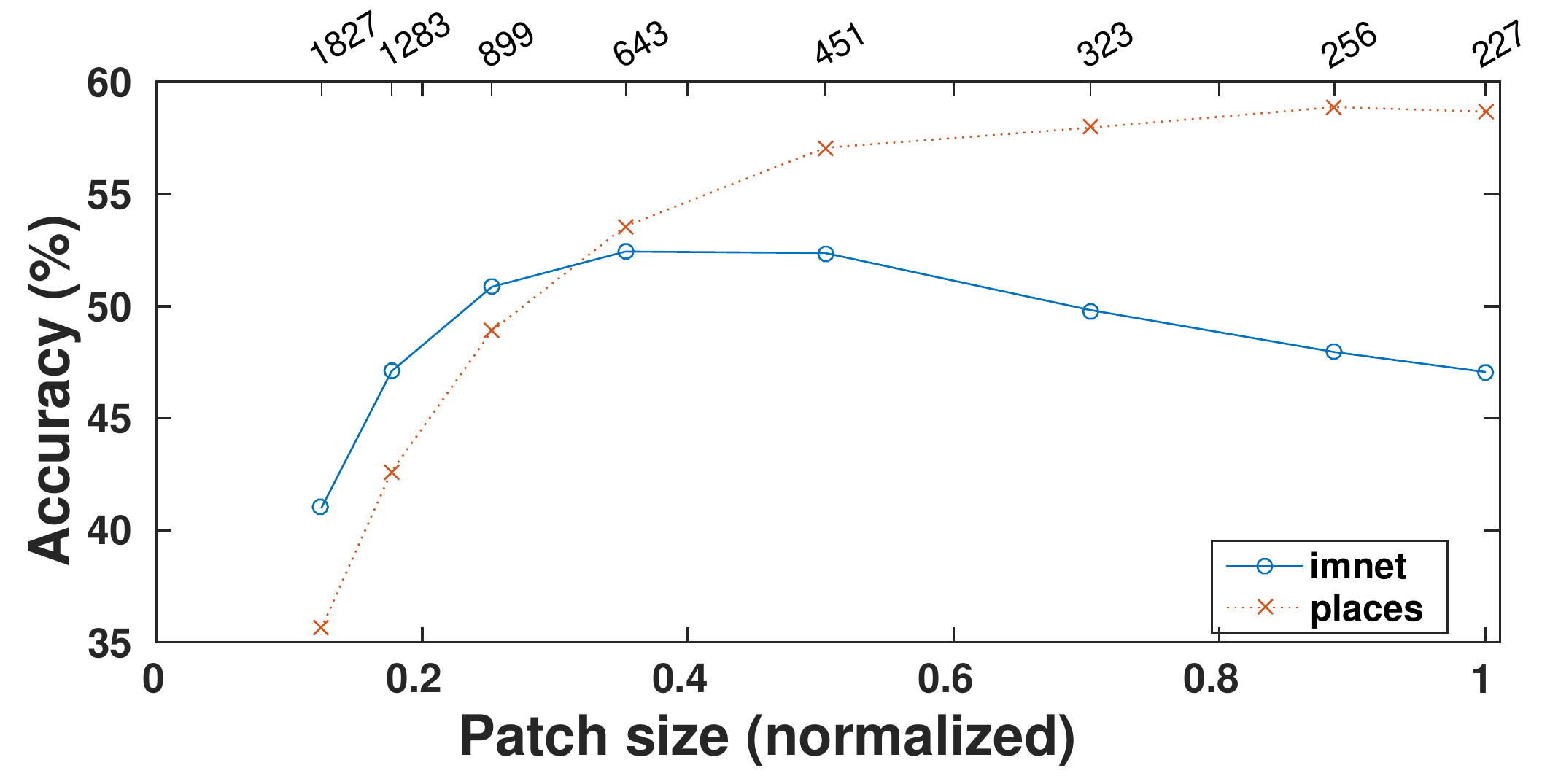}\tabularnewline
(c)\tabularnewline
\end{tabular}\caption{\label{fig:single_scale_accuracy}Scene recognition accuracy for different
scales: (a) 15 scenes, (b) MIT Indoor 67, and (c) SUN397.}
\vspace{-0.3cm}
\end{figure}

Average accuracy is a reasonable metric to evaluate a deep representation
in the context of a classification task. For the different scales,
we extracted \textit{fc7} activations locally in pacthes as features,
and then trained SVMs. In addition to the seven scales evaluated,
we included 256x256 pixels as a baseline, since off-the-shelf ImageNet-CNN
and Places-CNN are trained on this scale. The results for the three
datasets are shown in Fig.~\ref{fig:single_scale_accuracy}, with
similar patterns. Places-CNN achieves the best performance when is
applied globally at scene level (227x227 or 256x256), while rapidly
degrades for more local scales. ImageNet-CNN exhibits a very different
behaviour, with a more modest performance at global scales, and achieving
optimal performance on patches at intermediate scales, and outperforming
Places-CNN at most local scales. These curves somewhat represent the
operational curve of CNNs and the scale. In particular, the performance
of ImageNet-CNN can be increases notably just by using an appropriate
scale.

An interesting observation is that there is one point (around 643
or scale 0.35) that splits the range into two parts, one dominated
by ImageNet-CNN and another one dominated by Places-CNN, which we
can loosely identify as object range and scene range. We will use
this observation later in Section~4.4 to design spliced architectures.

\subsubsection{Effect of fine tuning}

A popular way to deal with dataset bias in CNNs is fine tuning, which
basically continues training on a pretrained model with the target
dataset data. Similarly in our case, we expect that fine tuning can
modify somehow the weights and thus adapt to the objects or at least
the scales in the target dataset. However, in practice that is often
not possible because of the limited data, overfitting and difficulty
of setting the training process itself. In our case, fine tuning on
scales where patches are very local is very difficult due since the
patch often contains objects or parts while labels indicates scenes
categories. In addition, the number of patches is huge, so only a
tiny fraction of them can be used in practice, rendering fine tuning
not very effective.

We evaluated fine tuning on MIT Indoor 67. For scales with few patches,
and thus limited training data, we only fine tune the fully connected
layers. For larger images we can collect more patches, up to 500K
patches (randomly selected). Fig.~\ref{fig:single_scale_accuracy}b
shows the results. Interestingly, there is a moderate gain in those
range of scales where the original CNNs perform poorly, i.e. global
scales for ImageNet-CNN and local scales for Places-CNN, while marginal
or no gain in ranges where they have already good performance. Thus,
fine tuning has certain ``equalizing'' effect over the accuracy
vs scale curve. but limited overall improvement. In particular the
gain is such that now Places-CNN (fine tuned) has the best performance
in the whole range of scales.

Fine tuning has impact mostly on the top layers, obtaining a similar
effect to adding a BOW pooling layer. However, the effectiveness is
limited, since intermediate layers remain biased to the (pre-)training
data.

\subsubsection{Discriminability and redundancy}

Accuracy provides a good indirect measure of the utility of the feature
for a given target task (e.g. scene recognition) via a classifier
(e.g. SVM). Here we also consider two information theoretic metrics
measuring directly the discriminability and redundancy of the deep
feature\cite{Peng2005}. We define the \textit{discriminability} of
a feature $\mathbf{x}=\left(x_{1,\cdots,}x_{4096}\right)$ with respect
to a set of classes $C=\left\{ 1,\cdots,M\right\} $ 
\[
D\left(\mathbf{x},C\right)=\frac{1}{\left|C\right|\left|S\right|}\sum_{c\in C}\sum_{x_{i}\in\mathbf{x}}I\left(x_{i};c\right)
\]
where $I\left(x_{i};c\right)$ is the filter $x_{i}$ and the class
$c$. In order to evaluate how redundant is the feature (compared
with other filters), we use the \textit{redundancy} of a feature $\mathbf{x}$,
defined as
\[
R\left(\mathbf{x}\right)=\frac{1}{\left|S\right|^{2}}\sum_{x_{j}\in\mathbf{x}}\sum_{x_{i}\in\mathbf{x}}I\left(x_{i};x_{j}\right)
\]

\begin{figure}
\centering{}\setlength{\tabcolsep}{0pt}
\renewcommand{\arraystretch}{1}%
\begin{tabular}{c}
\includegraphics[width=0.85\columnwidth]{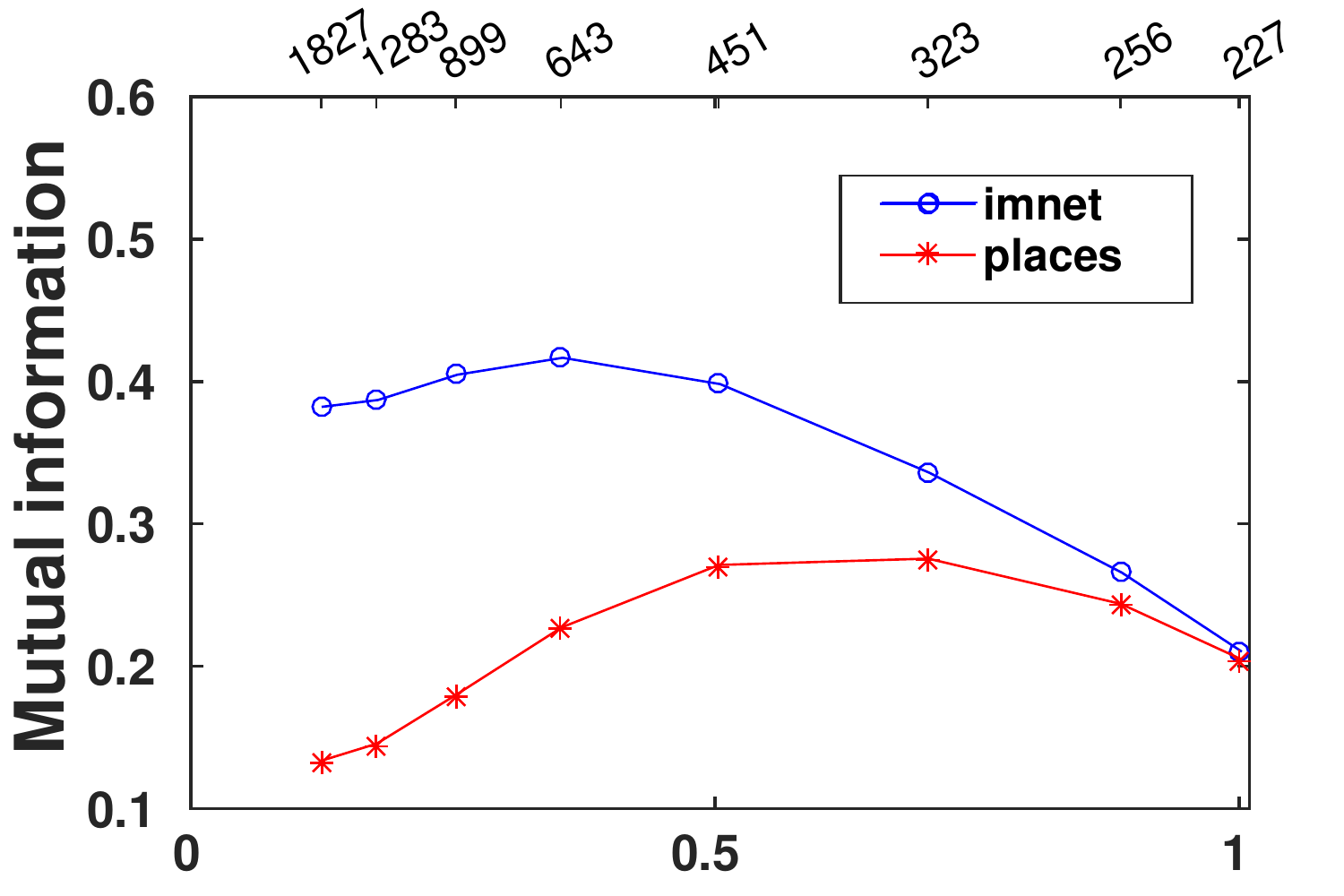}\tabularnewline
(a)\tabularnewline
\includegraphics[width=0.85\columnwidth]{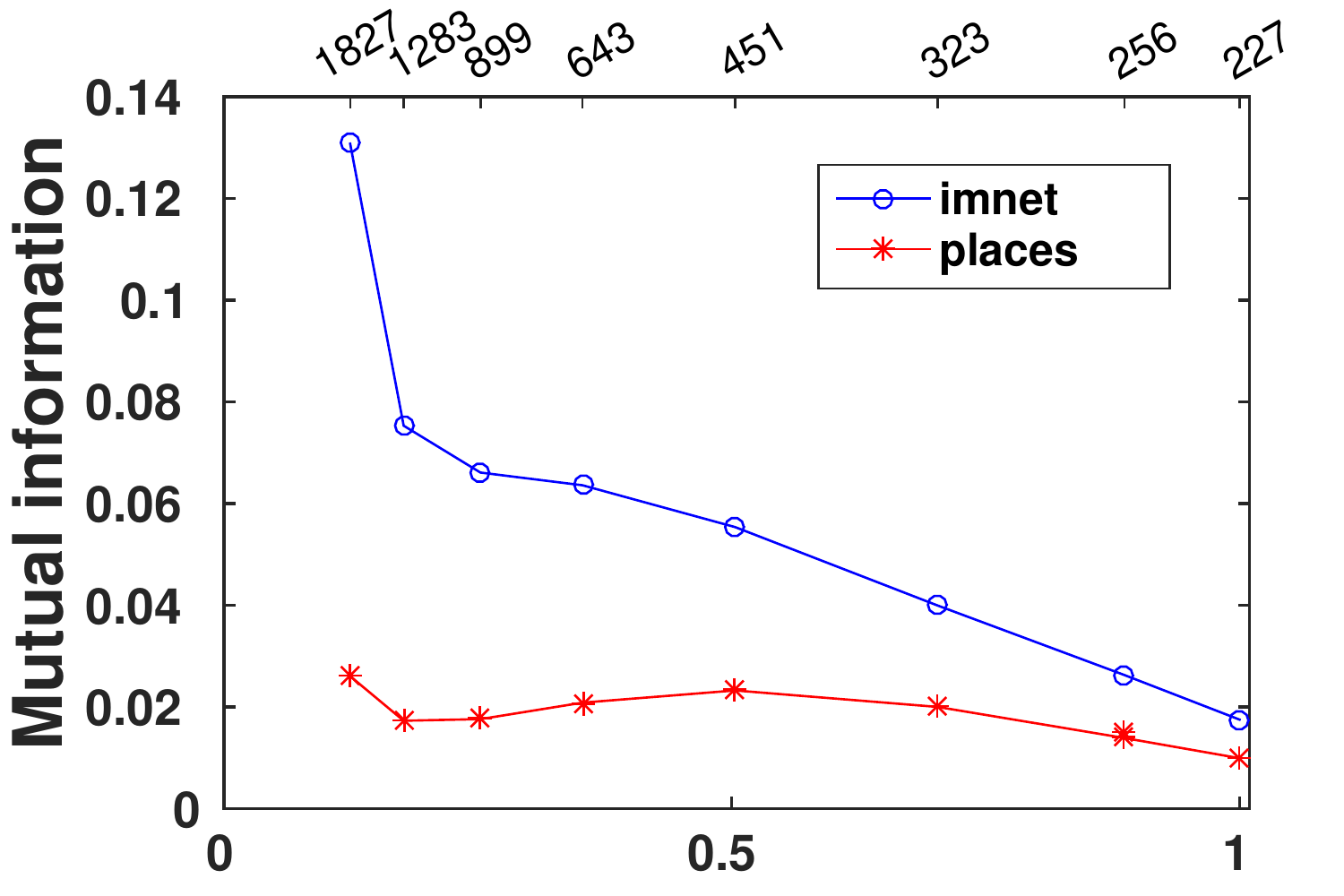}\tabularnewline
(b)\tabularnewline
\end{tabular}\caption{\label{fig:single_scale_mutual_information-1}Discriminability (a)
and redundancy (b) of fc7 feature in MIT Indoor 67.}
\vspace{-0.3cm}
\end{figure}

In the next experiment we compute $D\left(\mathbf{x},C\right)$ and
$R\left(\mathbf{x}\right)$ of the fc7 activation for ImageNet-CNN
and Places-CNN in MIT Indoor 67. While we can find similarities with
the accuracy curve, a direct comparison is not easy, since more discriminability
not always means higher accuracy. If we observe the discriminability
of ImageNet-CNN (see Fig.~\ref{fig:single_scale_mutual_information-1}a),
the curve follows a similar pattern to the accuracy, with a peak around
the scales where the accuracy was best, and bad discriminability at
global scales. Places-CNN extracts the most discriminative features
at more global scales. Comparing ImageNet-CNN and Places-CNN, the
former obtains more discriminative features yet also more redundant.
Too local scales (e.g. 1827x1827) increase significantly the redundancy
of the feature and the noise

\subsection{Two scales}

\begin{figure*}
\begin{centering}
\begin{tabular}{cc}
\includegraphics[bb=0bp 0bp 324bp 144bp,clip,width=0.49\textwidth]{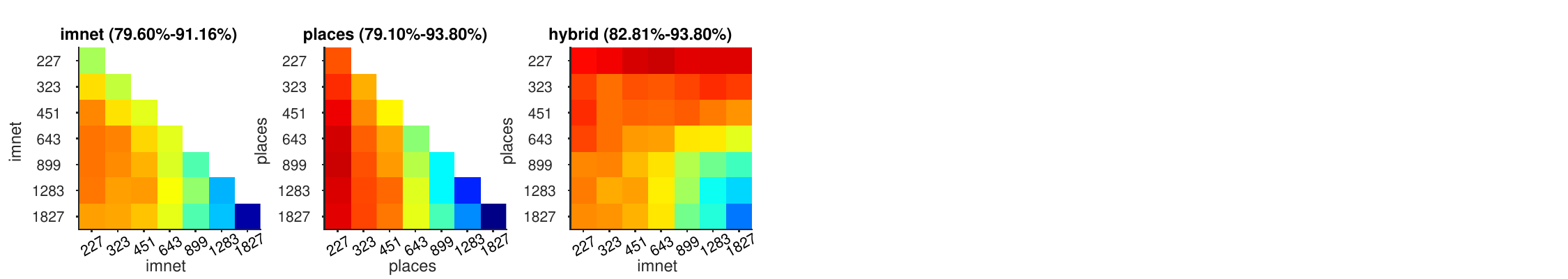} & \includegraphics[bb=0bp 0bp 324bp 144bp,clip,width=0.49\textwidth]{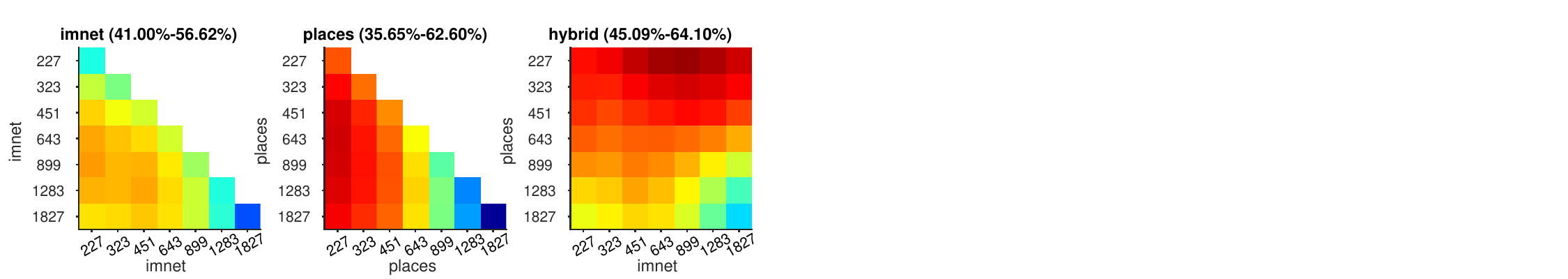}\tabularnewline
(a) & (b)\tabularnewline
\multicolumn{2}{c}{\includegraphics[clip,width=0.98\textwidth]{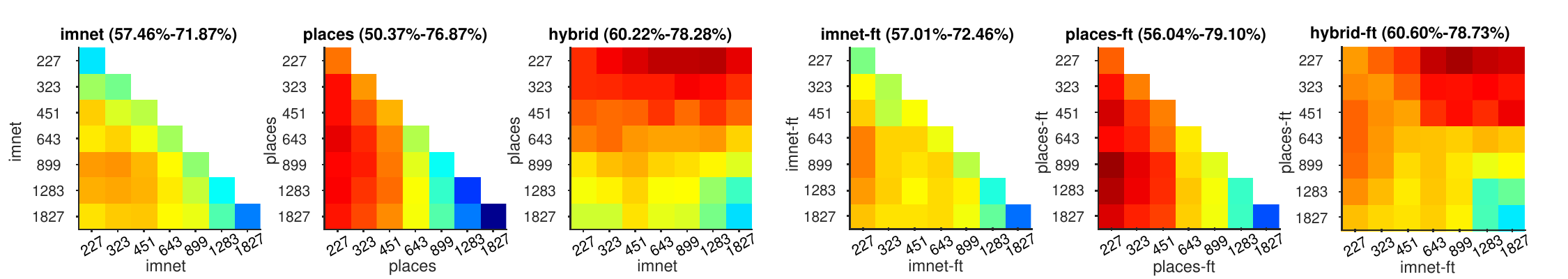}}\tabularnewline
\multicolumn{2}{c}{(c)}\tabularnewline
\end{tabular}\caption{\label{fig:dual_architectures}Accuracy in dual architectures combining
two networks (only ImageNet-CNNs, only ImageNet-CNNs and hybrid combinations):
(a) 15 scenes, (b) SUN397 and, (c) MIT Indoor 67 (and fine tuned versions).
Diagonals of only ImageNet-CNNs and only Places-CNNs variations show
single scale accuracy.}
\par\end{centering}
\centering{}\vspace{-0.3cm}
\end{figure*}

In the next experiment we evaluated pairwise combinations of CNNs
used at different scales. This dual architecture consists simply of
two CNNs processing images at different scales. We then concatenate
the two resulting fc7 activations into a 8192-dim feature and then
train the SVM. The results in Fig.~\ref{fig:dual_architectures}
show that the dual architectures with best performance are hybrid
combinations of Places-CNN extracting features at global scales (typically
227x227) with ImageNet-CNN extracting features from patches at more
local scales. The result is a considerable boost in the performance,
achieving a remarkable accuracy of 64.10\% on SUN397 using only two
AlexNet CNNs.

Note that this way of combining ImageNet and Places data is much more
effective than Hybrid-CNN\cite{Zhou2014b} (see Table~\ref{tab:summary_accuracy}).
Our dual architecture does not mix object and scene knowledge (obtained
from ImageNet and Places, respectively) and adapts the learned models
to scales with similar properties. Dixit et al\cite{Dixit2015} combine
Places-CNN with a four-scales architecture built on top of ImageNet-CNN.
Similarly to our framework, Places-CNN operates at scene scale while
ImageNet-CNN operates at object scales. Note, however, that we obtain
comparable performance on MIT Indoor 67 and significantly better on
SUN397, using just two networks instead of five.

\subsection{Multiple scales}

Finally, we consider the combination of all the scales to see whether
more complex architectures could be helpful in terms of accuracy.
In this experiment we evaluate the concatenation of all the fc7 features
of each of the seven scale-specific networks. In this case we use
PCA to reduce the dimension of each features vector so the combined
dimension is approximately 4096. We achieve 74.33\% (all scales using
ImageNet-CNN) and 78.21\% (all scales using Places-CNN) accuracy for
MIT Indoor 67, and 58.71\% and 63.81\% for SUN397, respectively. Note
that both are better than the corresponding dual architecture, yet
below the corresponding dual hybrids (78.28\% and 64.10\%). This suggests
than including more scales while keeping the same CNN model is marginally
helpful and increases significantly the extraction cost and the noise
in the representation.

\begin{figure}
\begin{centering}
\setlength{\tabcolsep}{0pt}
\renewcommand{\arraystretch}{1}\includegraphics[width=0.98\columnwidth]{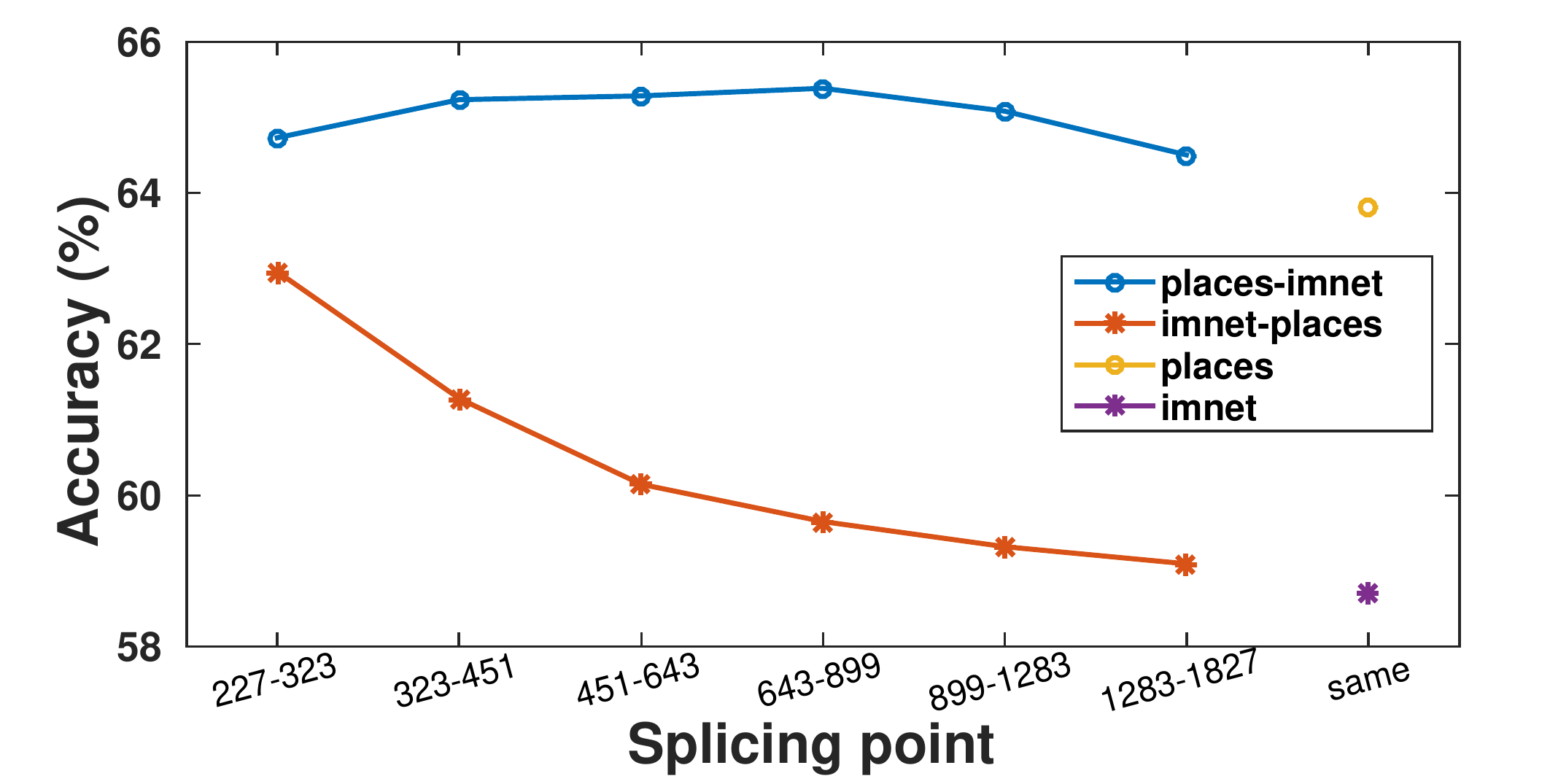}\caption{\label{fig:exp_full_and_splices}Accuracy on SUN397 of full and hybrid
spliced architectures (7 AlexNet networks). The combination \textit{same}
indicates that the 7 networks share the same CNN model (i.e. trained
with the same dataset).}
\par\end{centering}
\centering{}\vspace{-0.3cm}
\end{figure}

So the key is to find an appropriate combination of Places-CNNs and
ImageNet-CNNs. While in dual architectures evaluating all the combinations
is very costly, with seven networks the combinations is impractical.
Since the optimal ranges of both are complementary, we can design
the full hybrid architecture as global scales using Places-CNN and
local scales using ImageNet-CNN, just as shown in Fig.~\ref{fig:multiscale_representation}.
We can consider only one free parameter which is the splicing point.
The results for SUN397 are shown in Fig.~\ref{fig:exp_full_and_splices}.
As expected, we can reach slightly better performance (80.97\% and
65.38\%) than in dual architectures. The performance of hybrid spliced
is also significantly better than a 7 network architecture with a
fixed CNN model.

Finally we also evaluate a double full architecture, in which both
full ImageNet-CNN and full Places-CNN are combined in a complex 14
CNNs architecture by concatenating the previous features. This combination
does not help in MIT Indoor 67, and slightly in SUN397, reaching an
accuracy of 66.26\%.

\begin{table*}
\caption{\label{tab:summary_accuracy}Accuracy for different multi-scale variations
and architectures.}
\centering{}\begin{center}
\begin{threeparttable}%
\begin{tabular}{ccccccccc}
\hline 
\multirow{2}{*}{Architecture} & \multirow{2}{*}{Pretraining dataset} & \multirow{2}{*}{\#scales} & \multicolumn{2}{c}{15 scenes} & \multicolumn{2}{c}{MIT Indoor 67 (w/ FT)} & \multicolumn{2}{c}{SUN 397}\tabularnewline
 &  &  & Alex & VGG & Alex & VGG & Alex & VGG{\scriptsize{}\tnote{3}}\tabularnewline
\hline 
\hline 
\multirow{2}{*}{Baseline{\scriptsize{}\tnote{1}}} & IN & 1 & 87.60 & 90.69 & 61.49 & 72.31 & 47.93 & 55.19\tabularnewline
 & PL & 1 & 91.16 & 92.90 & 74.18 & 80.45 & 58.87 & 66.50\tabularnewline
\hline 
\multirow{2}{*}{Best single{\scriptsize{}\tnote{2}}} & IN & 1 & 88.54 & 91.86 & 66.64 (68.21) & 76.42 & 52.42 & 59.71\tabularnewline
 & PL & 1 & 91.65 & 93.73 & 72.76 (73.35) & 80.90 & 58.88 & 66.23\tabularnewline
\hline 
\multirow{2}{*}{Dual} & IN & 2 & 91.16 & 93.84 & 71.87 (72.46) & 79.04 & 56.62 & 61.07\tabularnewline
 & PL & 2 & 93.80 & \textbf{95.18} & 76.87 (79.40) & 83.43 & 62.60 & 68.49\tabularnewline
Dual hybrid & IN/PL & 1+1 & 93.80 & \textbf{95.18} & 78.28 (78.81) & 85.59 & 64.10 & 69.20\tabularnewline
Three{\scriptsize{}\tnote{4}} & IN/PL & 1+2 & 93.37 & 95.14 & 78.28 & \textbf{86.04} & 63.03 & \textbf{70.17}\tabularnewline
\hline 
\multirow{2}{*}{Full} & IN & 7 & 91.66 & 92.86 & 74.33 (75.97) & 70.22 & 58.71 & 55.18\tabularnewline
 & PL & 7 & 93.77 & 94.51 & 78.21 (79.70) & 77.81 & 63.81 & 58.80\tabularnewline
Full hybrid (spliced) & IN/PL & 7 & 93.90 & 94.08 & \textbf{80.97} (80.75) & 80.22 & 65.38 & 63.19\tabularnewline
Double full hybrid & IN/PL & 2x7 & \textbf{94.51} & 94.84 & \textbf{80.97} (79.85) & 80.7 & \textbf{66.26} & 62.01\tabularnewline
\hline 
\hline 
Hybrid-CNN\cite{Zhou2014b} & IN+PL & 1 & 53.86 & - & 70.80 & - & 53.86 & -\tabularnewline
MOP-CNN\cite{Gong2014} & IN & 3 & - & - & 68.88 & - & 51.98 & -\tabularnewline
MPP\cite{Yoo2015} & IN & 7 & - & - & 75.67 & - & - & -\tabularnewline
MPP+DSFL\cite{Yoo2015} & IN & 7+DSFL & - & - & 80.78 & - & - & -\tabularnewline
SFV\cite{Dixit2015} & IN & 4 & - & - & 72.86 & - & 54.4 & -\tabularnewline
SFV+Places\cite{Dixit2015} & IN/PL & 4+1 & - & - & 79.0 & - & 61.72 & -\tabularnewline
MetaObject-CNN\cite{Wu2015} & Hybrid (IN+PL)\cite{Zhou2014b} & 1 (variable) & - & - & 78.90 & - & 58.11 & -\tabularnewline
DAG-CNN\cite{Yang2015} & IN & 1 & - & 92.9 & - & 77.5 & - & 56.2\tabularnewline
DSP\cite{Gao2015,Wei2015} & IN & 1 & - & 91.78 & - & 78.28 & - & 59.78\tabularnewline
\hline 
Human (good)\cite{Xiao2014a} &  &  & \multicolumn{2}{c}{-} & \multicolumn{2}{c}{-} & \multicolumn{2}{c}{68.5\%}\tabularnewline
Human (expert)\cite{Xiao2014a} &  &  & \multicolumn{2}{c}{-} & \multicolumn{2}{c}{-} & \multicolumn{2}{c}{70.6\%}\tabularnewline
\hline 
\end{tabular}\begin{tablenotes}
\item[1] 256x256 central crop (conventional settings for single crop).
\item[2] Excluding 256x256.
\item[3] Six scales (1827x1827 was not included).
\item[4] Only evaluated the combination Places-CNN 227x227, Places-CNN 451x451, ImageNet-CNN 899x899.
\end{tablenotes}
\end{threeparttable}
\vspace{-0.3cm}
\end{center}
\end{table*}

\subsection{Deeper networks and other works}

The experiments presented so far are based on the AlexNet architecture.
Deeper architectures such as GoogLeNet\cite{Szegedy2015} and VGG-net\cite{Simonyan2015}
have demonstrated superior performance by exploiting deeper models.
We repeated some of the experiments using the 16 layer VGG architecture,
obtaining state-of-the-art results in the three datasets. The experiments
with VGG in dual architectures are consistent with those with AlexNet,
but with a more moderate gain. However, experiments combining more
networks were surprisingly disappointing, performing even worse than
single network baselines. VGG applied on small patches tends to be
very noisy with poor performance. We tried an intermediate hybrid
architecture, including a total of three scales, achieving slightly
better performance than with dual architectures.

Overall, for the small 15 scenes dataset, it seems that the performance
is somewhat saturated, with a best performance of 95.18\% (94.51\%
with AlexNet). The best performance in MIT Indoor 67 is 86.04\% (compared
with 80.97\% with AlexNet) and in SUN397 is 70.17\% (compared with
66.26\% with AlexNet). This performance is better than human recognition
by ``good workers'' (68.5\%), and close to human expert performance
(70.6\%), as reported in \cite{Xiao2010a}.

\section{Conclusions}

In contrast to previous works, in this paper we analyzed multi-scale
CNN architectures focusing on the local CNN model, rather than on
the pooling method. In particular, we showed that scaling images induces
a bias between training and test data, which has a significant impact
on the recognition performance. We also showed how ImageNet-CNN and
Places-CNN in this context are implicitly tuned for different scale
ranges (object scales and scene scales). Based on these findings,
we suggest that addressing this bias is critical to improve scene
recognition, and propose including scale-specific networks in the
multi-scale architecture. The proposed method is also a more principled
way to combine scene-centric knowledge (Places) and object-centric
knowledge (ImageNet) than previous attempts (e.g. Hybrid-CNN).

In fact, recent scene recognition approaches fall into two apparently
opposite directions: global holistic recognition (Places-CNN) versus
local object recognition and pooling (multi-scale CNNs). In this paper
we describe them as two particular cases in a more general view of
how multi-scale features can be combined for scene recognition. They
are not incompatible, and actually when combined properly to reduce
the dataset bias the results can be excellent, even reaching human
recognition performance simply with just two or three networks carefully
chosen. Our hybrid parallel architecture also suggests some similarities
with perceptual and cognitive models, where object recognition and
global scene features follow two distinct yet complementary neural
pathways which are later integrated to accurately recognize the visual
scene\cite{Oliva2014}.\\

\textbf{Acknowledgements. }This work was supported in part by the
National Basic Research Program of China (973 Program): 2012CB316400,
the National Natural Science Foundation of China: 61532018, 61322212,
51550110505 and 61550110505, the National Hi-Tech Development Program
(863 Program) of China: 2014AA015202, and the CAS President's International
Fellowship Initiative: 2011Y1GB05. This work is also funded by Lenovo
Outstanding Young Scientists Program (LOYS).

{\small{}\bibliographystyle{ieee}
\bibliography{osun_cvpr2016_v3}
 }{\small \par}
\end{document}